\documentclass{IEEEcsmag}

\usepackage[colorlinks,urlcolor=blue,linkcolor=blue,citecolor=blue]{hyperref}
\expandafter\def\expandafter\UrlBreaks\expandafter{\UrlBreaks\do\/\do\*\do\-\do\~\do\'\do\"\do\-}
\usepackage{upmath,color}
\usepackage{cite}

\jname{Computer}
\pubyear{2024}

\begin{document}

\sptitle{\normalsize This article has been accepted for publication in Computer. This is the accepted version which has not been fully edited and content may change prior to final publication. Citation information: DOI 10.1109/MC.2023.3327330}

\title{Understanding the Impact of Artificial Intelligence in Academic Writing: Metadata to the Rescue}

\author{Javier Conde}
\affil{Universidad Polit\'ecnica de Madrid, 28040 Madrid, Spain}

\author{Pedro Reviriego}
\affil{Universidad Polit\'ecnica de Madrid, 28040 Madrid, Spain}

\author{Joaqu\'in Salvach\'ua}
\affil{Universidad Polit\'ecnica de Madrid, 28040 Madrid, Spain}

\author{Gonzalo Mart\'inez}
\affil{Universidad Carlos III de Madrid, 28911 Madrid, Spain}

\author{Jos\'e Alberto Hern\'andez}
\affil{Universidad Carlos III de Madrid, 28911 Madrid, Spain}

\author{Fabrizio Lombardi}
\affil{Northeastern University, Boston, MA 02115, USA}


\begin{abstract}\looseness-1 The development of generative AI tools that can create diverse content such as text and images, is poised to have a great impact on academic writing. Understanding this phenomenon is far from trivial and needs to be carefully studied. The first step is to be able to identify papers written with the help of AI, ideally knowing which tools have been used, in which parts of the paper, and how they have been used. This  enables tracking the number of papers that use AI tools as well as the list of employed AI tools but it  will also make possible the generation of large datasets of documents that have used different AI tools. These datasets are needed to understand the impact of AI tools on academic language and their potential effects in the long term. Unfortunately, getting the details of the use of AI in academic papers does not seem to be possible unless AI-related metadata is defined with a common format and added to the papers. This column advocates the implementation of such metadata and discusses the potential benefits of having metadata in academic publications

\end{abstract}

\maketitle

\chapteri{T}he rapid development and exponential adoption of artificial intelligence (AI) tools that can generate text for different tasks, are poised to have significant and far reaching implications in many sectors  \cite{ChatGPTOverview}. These tools can for example summarize, translate, or paraphrase text, but also can write text on any topic and provide relevant citations \cite{ChatGPTOverview2}. These capabilities are useful for academic writing and many researchers are starting to use AI tools as assistants when writing papers \cite{ChatGPTAcademic1}. The use of AI tools poses many potential issues, ranging from the accuracy of the text generated that may contain false statements to ethical concerns \cite{ChatGPTAcademic2}. Therefore, the impact of AI tools should be carefully analyzed, not only in research datasets but also in real data as AI-generated content starts to become widespread and for academic writing in published papers.

There are many angles to be considered when analyzing the impact of AI tools in academic writing. For example, will the use of AI tools have an impact on the citations of the papers? Which authors are more prone to use those tools? Which tools are more popular and for which tasks? Those are some of the basic questions but more fundamental aspects have to be considered. Initial studies suggest that AI tools will have an impact on the language itself \cite{HCC1, HCC2}. So for example, will linguistic features of papers written with the assistance of AI tools be different from those of human-written papers?  Will the use of AI tools introduce biases in the vocabulary \cite{reviriego2023playing} used? The questions are numerous and soon academic papers written with the help of AI tools will become common. The answers to the questions lie on those papers. 

To analyze the impacts of these AI tools, the first step is to reliably identify AI-generated content with as many details as possible; for example, knowing the AI tool, its version, and the tasks for which it was used to assist in the writing. Unfortunately, such information is generally not available in academic papers. A potential solution is to use tools designed to detect AI-generated text \cite{ChatGPTDetection}. This however has many limitations because these tools have limited accuracy and have to be constantly evolving to keep track of new AI-generative tools and features. Even if accurate, such tools can only provide limited information, but not the details on the AI tool and version used. Another possible direction is the policies being implemented by many publishers, for example, the IEEE, to request authors to disclose the use of AI tools in the acknowledgment section. However, this approach also has limitations because there is no standard form to report the use of AI tools and provide additional details. Finally, for both approaches, even if the information could be extracted reliably (which is not the case), gathering even basic information such as the percentage of papers that have used a given tool requires checking the full text of all papers. This is clearly not efficient.       

A more scalable and future-proof solution is  to add metadata describing the use of AI tools in each academic paper. This enables queries on the metadata on large numbers of documents but also the definition of common metadata formats that enable interoperability among different publishers. For example, the addition of the tool, version, and task makes answering the first set of simple questions discussed previously trivial. Interestingly, this metadata enables the generation of large corpora of text written with the assistance of AI tools from which the answers to the second set of questions can be extracted. The use of metadata also enables tracking the evolution of the use of AI tools, so making possible comparison among different tools, or for the same tool among different versions. Next, we discuss the metadata that should be added and illustrate the potential benefits.


\section{METADATA FOR THE USE OF AI IN ACADEMIC WRITING}

The information on the use of AI in an academic paper can be captured at different levels of detail, from a simple flag indicating the use of AI to assist the authors to detailed logs of the interactions with the AI tool. As it tends to happen, too little or too much detail is not good, so next, we try to list the main information that could be relevant to analyze the impact of AI in academic writing:   

\begin{enumerate}
    \item AI tool used. 
    \item Version of the tool.
    \item Main parameters of the tool.
    \item Use of the tool (translation, summarization, writing, citations, etc.). 
    \item Parts of the paper on which the tool was used.
\end{enumerate}

The information can be divided into three categories: which tool was used, how it was used, and where it was used. The knowledge of the tool enables a comparative analysis of different tools but also the study of the evolution of a given tool or its parameters. Similarly, knowing for what the tool was used provides information to analyze the impact of AI on different aspects of academic writing. Finally, knowing where it was used enables the extraction of the relevant sections of the paper for further analysis. The three categories are illustrated in Figure 1.

\begin{figure*}
\label{fig:WHW}
\centerline{\includegraphics[width=36pc]{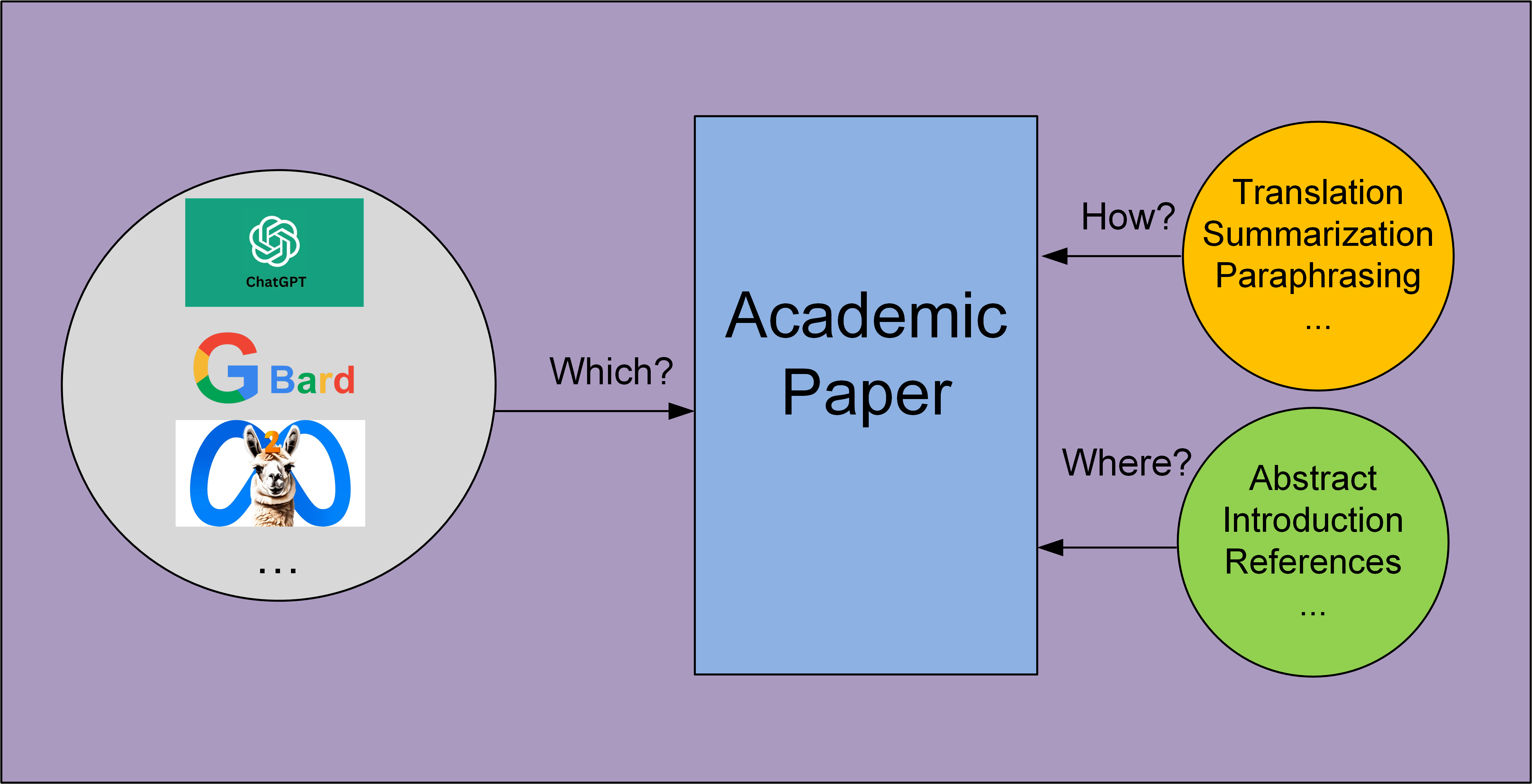}}
\caption{Relevant information (which, how and where) on the use of AI in an academic paper.}
\end{figure*}

\subsection{Which AI tools and parameters}

This group of metadata must capture the tool, version and configuration. The data can be coded in JSON, XML, or any other convenient format to automate its retrieval and processing. The following list is an example of how the data could look like:  

\begin{itemize}
    \item Name: GPT-4
    \item Description: GPT-4 is a large multimodal model (accepting text inputs and emitting text outputs today, with image inputs coming in the future) that can solve difficult problems with greater accuracy than any of our previous models, due to its broader general knowledge and advanced reasoning capabilities
    \item Version: gpt-4-0613
    \item URL: \url{https://platform.openai.com/docs/models/gpt-4}
    \item Author: OpenAI
    \item Web: \url{https://chat.openai.com//}
    \item Size: not disclosed
    \item Window: 8192 tokens    
    \item Temperature: 0
    \item Context: default
\end{itemize}

The metadata includes information on the AI model itself but also on the specific parameters' values selected when using it to write the paper. This enables the evaluation of the impact of the model parameters on the written text.

\subsection{Where AI was used}

The information on where AI tools were used can be described with the parts of the papers. For example the abstract, each of the sections, the figures, the tables or the references. This enables the identification of the text for which AI assistance has been used. 

\subsection{How AI was used}

AI tools can be used for many different tasks: summarizing, translation, paraphrasing, finding related work and citations, etc. So, it is important to have information on how AI tools were used in the paper. For example, we can encode in the metadata that GPT-4 (so the “which”) was used to summarize (the “how”) and write the abstract (the “where”).

\section{UNDERSTANDING THE IMPACT OF AI IN ACADEMIC WRITING}

Let us consider now that we have a large corpus of papers and we want to know how many of them have used AI to summarize the abstract. Without metadata, all papers look the same (Figure 2, left), so we have to extract the text and either try to detect the use of AI in the abstract or find a disclosure of the authors that states the use of AI in the abstract. Instead if the proposed metadata has been added to the paper, we can just look at the how (summarizing) and where (abstract) to find the papers. The papers are now marked and can be easily identified (Figure 2, right).

\begin{figure*}
\label{fig:Metadata}
\centerline{\includegraphics[width=40pc]{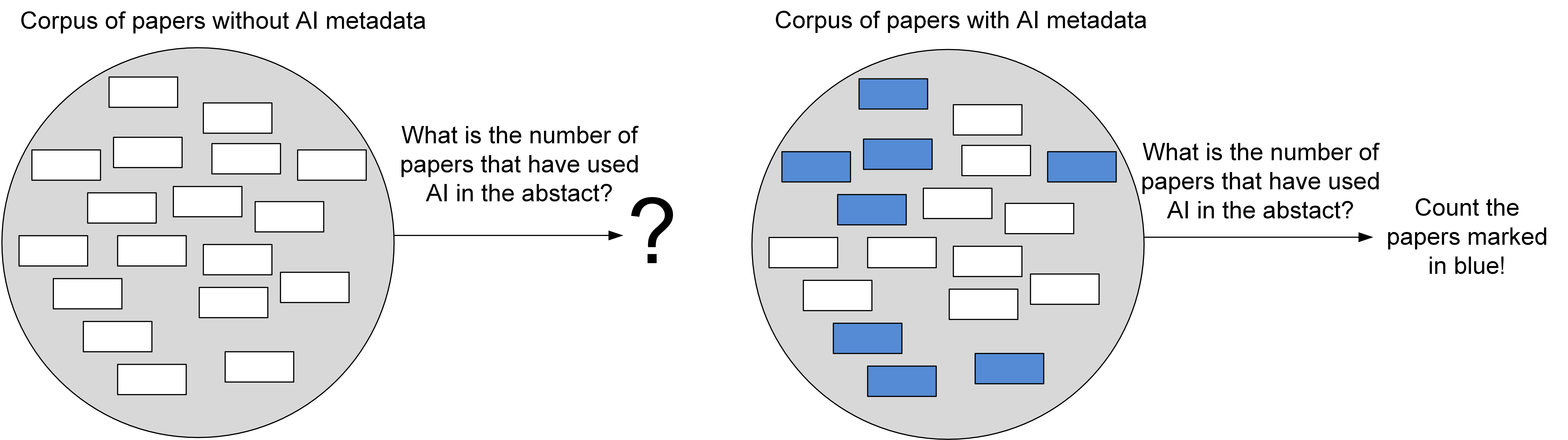}}
\caption{Finding papers with specific AI use features: (left) no metadata is used so all papers look the same and have to be parsed to try to extract the information; (right) metadata is used and papers can be easily identified without having to access the full text.}
\end{figure*}

\begin{figure*}
\label{fig:UseMetadata}
\centerline{\includegraphics[width=24pc]{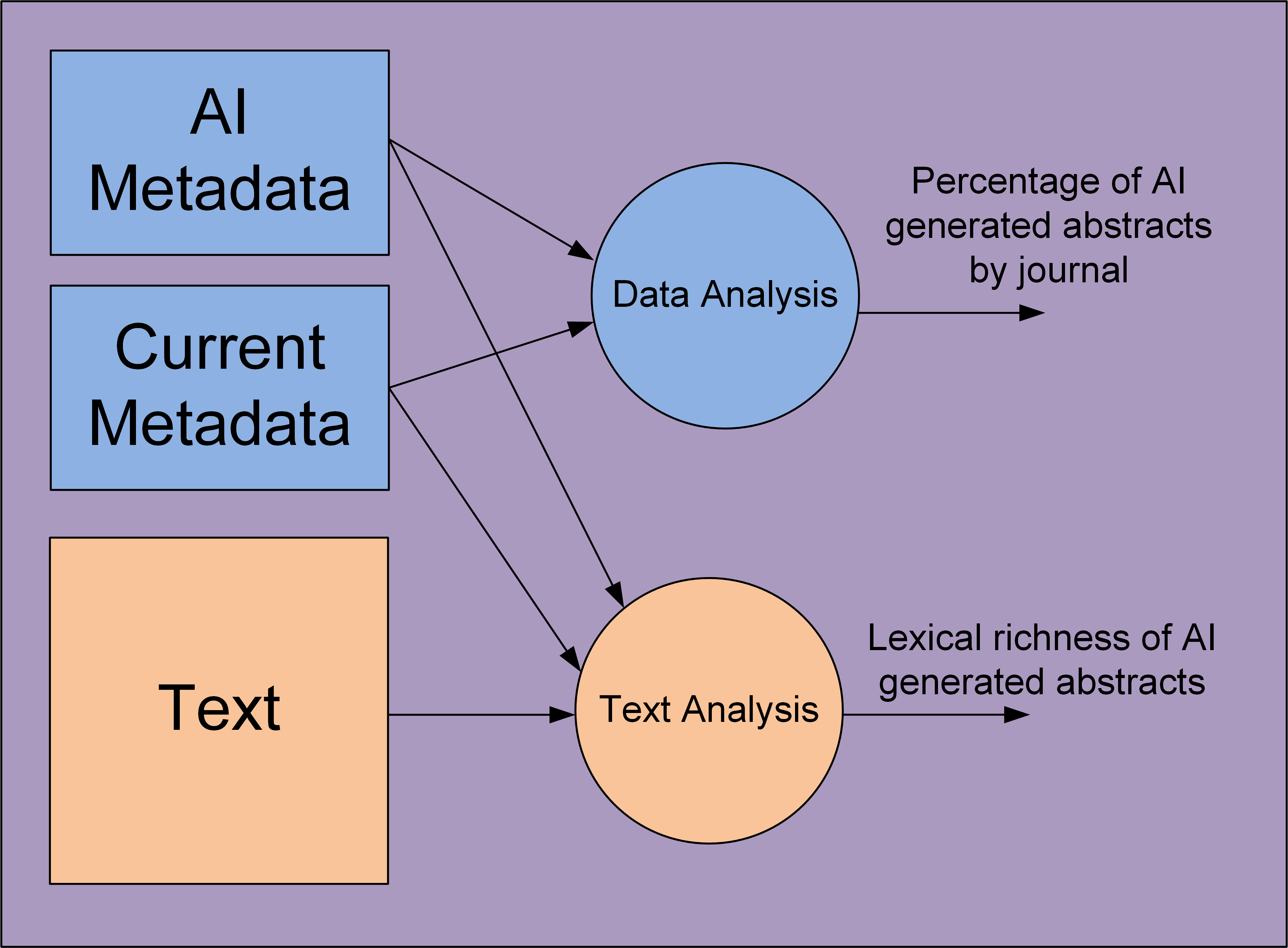}}
\caption{Use of AI metadata for data and text analysis.}
\end{figure*}

The metadata can be used to analyze many aspects of the use of AI in academic writing, for example, we can analyze: 

\begin{enumerate}
    \item The adoption of the different AI tools and their variations over time. 
    \item The tasks for which AI tools are more frequently used. 
    \item The parts of the papers for which AI tools are more widely used.
    \item The correlation between the use of AI in a paper and its citation and popularity metrics.
    \item The use of AI tools in the different journals and conferences by the same publisher.
    \item The use of AI tools by authors of different regions and institutions.
    \item The use of AI tools by an author over time.
\end{enumerate}

These are just few simple examples that show how adding simple metadata to academic publications makes it possible to gain different insights into the impact of AI tools in academic writing. These analyses can be easily run and automated once the metadata is integrated into the publisher's information systems opening the possibility of periodic reporting and analysis. 

The proposed metadata not only enables purely AI-related studies but also those that combine existing metadata such as for example the number of citations and the use of AI tools. The combinations of parameters that can be studied are endless, one possibility is to feed the raw into a machine learning system to extract patterns and relations.    

In addition to the analysis of the metadata, it is also possible to use the metadata to perform analysis of the content of the main body of the papers. This is of interest when studying the impact of AI tools on the linguistic features of the text generated \cite{HCC2}, for example, lexical richness \cite{reviriego2023playing}. Using the metadata, the text in papers (or sections within those papers) in which AI tools have been used, can be extracted to build a corpus of text generated for example with the assistance of a given tool. More generally corpora of text for a given set of AI metadata values can be easily generated. This is very interesting because it enables the analysis of the impact of AI tools on real data generated by many different users. The availability of such corpora opens new possibilities, such as for example:

\begin{enumerate}
    \item Analyze how AI tools affect the features of generated text depending on the tool and version.
    \item Analyze how AI tools affect the features of generated text depending on the task.
    \item Compare text generated by the same authors with and without the help of AI tools.
    \item Compare text generated by native and not native speakers as authors with and without AI tools.
    \item Provide datasets of user-generated data to develop and validate tools that detect AI-generated text. 
    \item Compare text generated by authors when using AI tools with text generated directly by AI tools to understand how the assistance rather than the direct use of AI tools impacts the text.
\end{enumerate}

The overall scenario is illustrated in Figure 3 with an example that shows how metadata can be used to understand the summarizing process with AI tools as used in different journals and also how metadata is combined with full text to analyze the lexical richness of AI generated abstracts.

\section{CONCLUSION}

This column proposes adding metadata on the use of artificial intelligence to scientific publications. The provision of having this metadata is critical for the analysis and understanding of artificial intelligence and its impact on academic writing. It is important to note that the implementation of this solution will require changes to scientific journal and academic database systems by adding new fields to store the new metadata.

\section{ACKNOWLEDGMENTS}
This work was supported by the Agencia Estatal de Investigaci\'on (AEI) (doi:10.13039/501100011033) under Grant FUN4DATE (PID2022-136684OB-C21/22).

\bibliographystyle{IEEEtran}
\bibliography{MetadataAI}

\begin{IEEEbiography}{Javier Conde}{\,} is an assistant professor at the Universidad Polit\'ecnica de Madrid, 28040 Madrid, Spain. Contact him at javier.conde.diaz@upm.es
\end{IEEEbiography}

\begin{IEEEbiography}{Pedro Reviriego}{\,} is an associate professor at the Universidad Polit\'ecnica de Madrid, 28040 Madrid, Spain. Contact him at pedro.reviriego@upm,es
\end{IEEEbiography}

\begin{IEEEbiography}{Joaqu\'in Salvach\'ua}{\,} is an associate professor at the Universidad Polit\'ecnica de Madrid, 28040 Madrid, Spain. Contact him at joaquin.salvachua@upm.es
\end{IEEEbiography}

\begin{IEEEbiography}{Gonzalo Mart\'inez}{\,} is a PhD student at the Universidad Carlos III de Madrid, 28911 Madrid, Spain. Contact him at gonzmart@pa.uc3m.es
\end{IEEEbiography}

\begin{IEEEbiography}{Jos\'e Alberto Hern\'andez}{\,} is an associate professor at the Universidad Carlos III de Madrid, 28911 Madrid, Spain. Contact him at jahgutie@it.uc3m.es
\end{IEEEbiography}

\begin{IEEEbiography}{Fabrizio Lombardi}{\,} is a professor at Northeastern University, Boston, MA 02115, USA. Contact him at lombardi@coe.northeastern.edu
\end{IEEEbiography}

\end{document}